\documentclass[lettersize,journal]{IEEEtran}
\usepackage{amsmath,amsfonts}
\usepackage{algorithmic}
\usepackage{algorithm}
\usepackage{array}
\usepackage{amssymb}
\usepackage{bbding}
\usepackage{graphicx}
\usepackage{parskip}
\usepackage{textcomp}
\usepackage{xcolor}
\usepackage{booktabs}  
\usepackage{threeparttable} 
\usepackage{textcomp}
\usepackage{stfloats}
\usepackage{url}
\usepackage{verbatim}
\usepackage{graphicx}
\usepackage{cite}
\usepackage{hyperref}
\usepackage{dsfont}
\usepackage{verbatim}
\usepackage{subfigure}
\usepackage{multicol}
\usepackage{multirow}
\usepackage{mathrsfs}
\usepackage{fourier} 
\usepackage{array}
\usepackage{makecell}

\usepackage{xspace}
\makeatletter
\DeclareRobustCommand\onedot{\futurelet\@let@token\@onedot}
\def\@onedot{\ifx\@let@token.\else.\null\fi\xspace}

\def\ie{\emph{i.e}\onedot}

\def\wrt{w.r.t\onedot} 

\makeatother

\begin{document}

\title{Fast Satellite Tensorial Radiance Field for Multi-date Satellite Imagery of Large Size}
\author{Tongtong Zhang, Yuanxiang Li*, ~\IEEEmembership{Member, ~IEEE} \thanks{Tongtong Zhang, Yuanxiang Li are with the Department of Information and Control, School of Aeronautics and Astronautics, Shanghai Jiao Tong University.}}



\maketitle
\begin{abstract} 
    Existing NeRF models for satellite images suffer from slow speeds, mandatory solar information as input, and limitations in handling large satellite images. In response, we present SatensoRF, which significantly accelerates the entire process while employing fewer parameters for satellite imagery of large size. Besides,  we observed that the prevalent assumption of Lambertian surfaces in neural radiance fields falls short for vegetative and aquatic elements.
    In contrast to the traditional hierarchical MLP-based scene representation, we have chosen a multiscale tensor decomposition approach for color, volume density, and auxiliary variables to model the lightfield with specular color. 
     Additionally, to rectify inconsistencies in multi-date imagery, we incorporate total variation loss to restore the density tensor field and treat the problem as a denosing task. 
    To validate our approach, we conducted assessments of SatensoRF using subsets from the spacenet multi-view dataset, which includes both multi-date and single-date multi-view RGB images.
 Our results clearly demonstrate that SatensoRF surpasses the state-of-the-art Sat-NeRF series in terms of novel view synthesis performance. Significantly, SatensoRF requires fewer parameters for training, resulting in faster training and inference speeds and reduced computational demands.
\end{abstract}
\begin{IEEEkeywords}
Neural Radiance Field, tensor decomposition, acceleration, DSM.
\end{IEEEkeywords}

\section{Introduction}
\label{sec:introduction}
\IEEEPARstart{T}{he} reconstruction of the earth's surface from high-resolution satellite imagery has consistently held significant importance across various domains. Conventional approaches within this domain typically tackle the challenge by solving multiple stereo problems to estimate disparity \cite{asp_2016,s2p_2017,micmac_2018,satcolmap_2019,gong2019dsm_2019}.
However, these approaches encounter inherent challenges during the dense matching phase. Despite the incorporation of deep learning techniques into satellite imagery, such as differentiable warping of Rational Polynomial Camera (RPC) for multi-view cost volume in \cite{satmvs_2023}, the computation and time costs are still substantial. Besides, most of the deep-learning based stereo problem require ground truth depth map as supervision \cite{deepsfm}. 
\begin{figure}
    \centering
    \includegraphics[width=9cm]{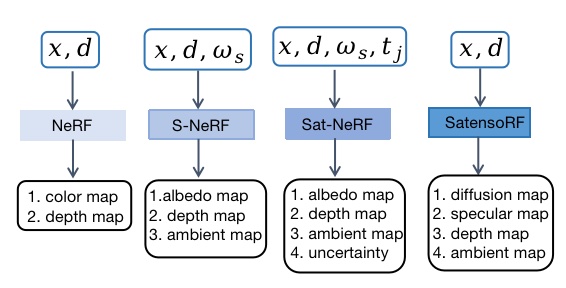}
    \caption{The paradigm comparison of SatensoRF and the previous satellite-related NeRFs. SatensoRF has no requirements on additional input other than position $x$ and viewing direction $d$, which is calculated from the sampled point cloud \wrt RPC.$\omega_s$ refers to solar direction and $t_j$ referes to the image index embeddings.
    }
    \label{fig:comp_paradim}
\end{figure}

\begin{figure}
    \centering
    \includegraphics[width=9cm]{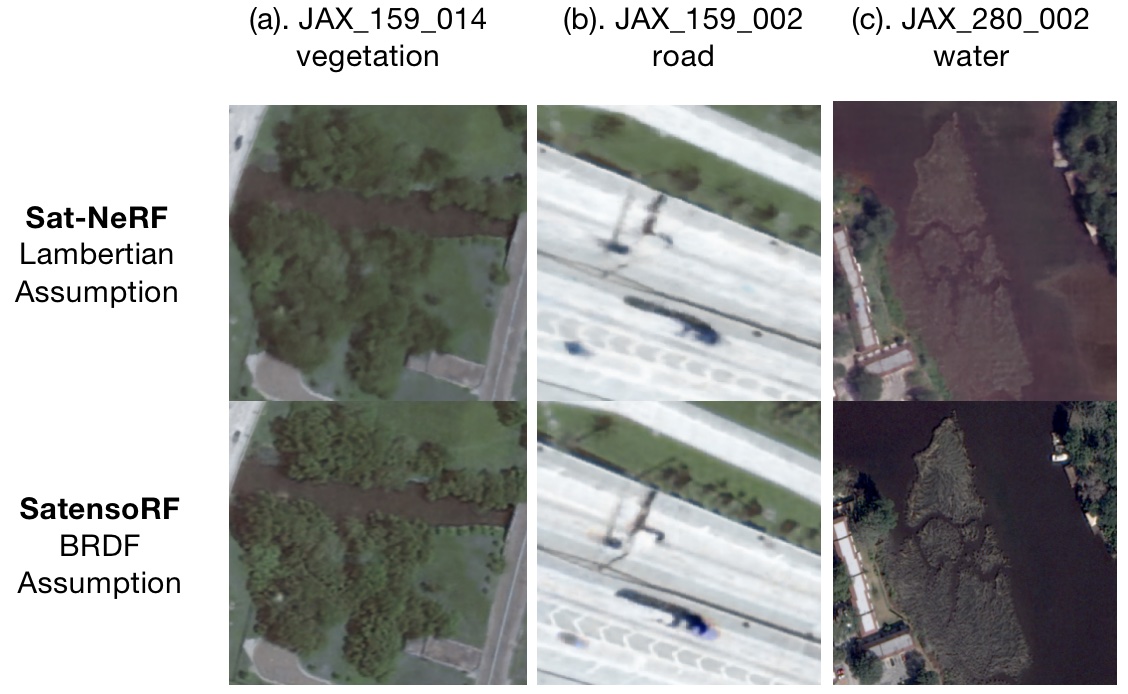}
    \caption{Rendering results on specific patches with different light field assumptions. (a). For vegetational regions, SatensoRF with BRDF assumptions yields better tree details. (b). For reflective roads, SatensoRF results in finer edges. (c). SatesoRF is more capable of capturing water regions. }
    \label{fig:comp_assumption}
\end{figure}
\begin{figure}
    \centering
    \includegraphics[width=8cm]{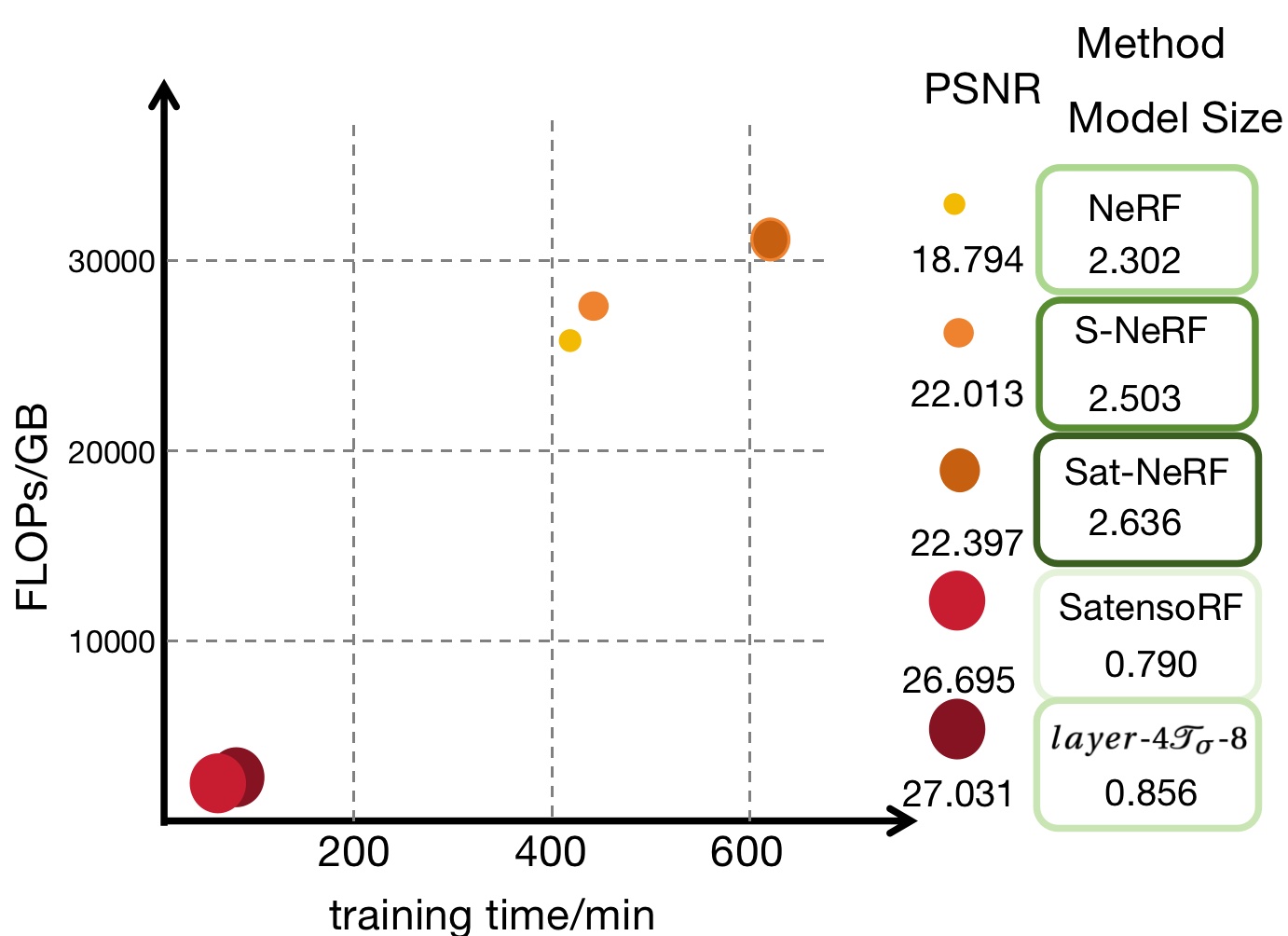}
    \caption{Comprehensive comparison of SatensoRF with SOTA satellite NeRFs, on subset 159 of spacenet multi-view dataset with uncropped size. SatensoRF achieves better rendering results with faster training speed with less computation cost, and $layer-4\mathcal{T}_{\sigma}-8$  refers to SatensoRF with 4 layers MLP and 8 levels of tensor radiance fields.}
    \label{fig:comp_all}
\end{figure}

In recent years, the emergence of the Neural Radiance Field (NeRF) \cite{nerfreview} has sparked a revolutionary shift, particularly within the fields of computer graphics and computer vision. 
Reconstruction methods based on NeRF has much less dependency on dense matching thus requiring no groundtruth depth map for supervision.
The fundamental concept behind the original NeRF is to use a simple, fully connected neural network to model a continuous scene. This network takes ray samples  and viewing directions as inputs and maps them to RGB color and volume density. Subsequently, differential volume rendering facilitates end-to-end training for scene representation. Subsequent adaptations of NeRF in the aerospace domain have yielded significant benefits for drone applications \cite{locnerf,meganerf}, as well as satellites \cite{snerf,satnerf}.
The common domain gap for migrating canonical RGB input to satellite images includes the following aspects:
(1). radiometric inconsistencies among images across multiple dates. 
(2). discrepancies in the ray modeling between perspective cameras and rational polynomial cameras (RPCs), which require distinct sampling strategies. 
For the first gap,
S-NeRF \cite{snerf} introduces a remodeled neural volume rendering approach based on solar irradiance and incorporates the solar directions $\omega_s$ as input, as depected in (b) of Fig.~\ref{fig:comp_paradim}. Building upon this, Sat-NeRF \cite{satnerf} extends the framework to model transient objects by regressing uncertainty to dynamically adjust the loss. 
Sat-NeRF also draws inspiration from NeRF in the wild (NeRF-w) \cite{nerfw}, by introducing image index embedding $t_j$  to supplement the multi-date difference in (c) of Fig.~\ref{fig:comp_paradim}.
For the second gap, Sat-NeRF also adapts the point sampling model to accommodate RPCs. 
However, despite the successful adaptation of neural radiance fields to multi-date satellite imagery, significant practical challenges persist.

The first left-over problem is that the conventional NeRF parameterization techniques in S-neRF and Sat-NeRF inherently retain the efficiency limitations of vanilla NeRF.
In particular, the two-stage coarse-to-fine training approach requires a wide array of images from distinct viewpoints to ensure the seamless synthesis of continuous views. This rendering configuration results in dense querying of Multi-Layer Perceptrons (MLPs) at all query points.
Besides, the purely MLP-based model introduces a trade-off between memory consumption and batch size, consequently impeding the overall  efficiency and constraining the input image size.

The second problem is that Sat-NeRF requires additional input such as sun azimuth, elevation, as shown in Fig.~\ref{fig:comp_paradim}. On the one hand, these additional input is essential for Sat-NeRF but might be unaccessible for some datasets. On the other hand, additional input tensors cause more computations, and further slower the running process.
 Specifically, we eliminate the following superfluous input variables:
\begin{enumerate}
    \item The solar direction in \cite{snerf,satnerf} used for forming the light field over the assumed Lambertian surface.
    \item The index embedding in \cite{nerfw,snerf,satnerf} to encode the transient objectives in multi-date photo collections.
\end{enumerate}

The third problem is the over-simplified light field assumption. S-NeRF and Sat-NeRF have operated under the assumption that the earth's surface follows strict Lambertian law. They break down the rendered color into distinct elements—surface albedo, ambient color, and leveraging factor—dependent on solar orientation. Nevertheless, real-world surfaces often deviate from Lambertian qualities, exhibiting bidirectional reflection traits to varying degrees \cite{nonlambertian2001}. This phenomenon applies to surfaces like bodies of water, vegetation, and asphalt roads, and the distinction is shown in Fig.\ref{fig:comp_assumption}. 

For the first issue, our proposal tensorizes the 3D scene using Level-of-Detail representations. The appearance field of the earth's surface can be decomposed into multiscale representations, manifested as 3D voxel grids with three channels. Similarly, the volume density field of the Earth's surface and the global hue bias are expressed as 3D voxel grids with a single channel.  SatensoRF largely improves the running speed and effectively reduces the total model size. Therein, SatensoRF requires less running memory and can be better applied to large satelite images than Sat-NeRF

For the second and third issue, we introduce a revised light field model with SatensoRF in Section \ref{sec:lightfield} to account for the BRDF properties, without need of additional solar information and index embeddings. In Fig.~\ref{fig:comp_assumption}, the proposed model encompassing the reflective light yields finer details of trees, metal surface of cars, and the water surface.

Moreover, regarding the cancelation of index embedding, we leverage the low-rank structure to handle the transient objects. This is achieved by treating transient objects as separate black-box noise components.
Consequently, the task at hand transforms into the restoration of a stable volume density field, as further elucidated in Section \ref{sec:total_variation}.
 Then, in virture of the sparsity of the tensor radiance field, we apply total variation regularization to reduce the noise, and then prevents the influence of the inconsistency among images from different dates. 

To sum up, our work makes several notable contributions:
\begin{itemize}
    \item Introducing a lightweight and expedited tensor radiance field framework that requires less training and inference time.
    \item Reducing model sizes, and overall computational memory cost, and demonstrating superior results on large-sized satelllite images, compared to previous NeRF-based approaches. 
    \item Capturing the inherent anisotropic scattering properties of the earth's surface without presuming the Lambertian law, achieved through a lightweight model design with improved performance.
    \item Significantly reducing input dependencies by enabling SatensoRF to operate without relying on solar direction information.
\end{itemize}
The proposed SatensoRF are evaluated on different areas of interest the multi-date WorldView-3 images of the original size.  For altitude estimation, we adopt a lidar digital surface model (DSM) of resolution 0.5 m/pixel as evaluation reference. SatensoRF is compared to the state-of-the-art NeRF, S-NeRF, and Sat-NeRF. 

\section{Related Work}
\subsection{Acceleration Framework for Neural Radiance Fields}
In the conventional NeRF approach, the rendering step involves the dense evaluation of MLPs at all query points along each camera ray during numerical integration, leading to slow speed and extensive computational demands. To address this, baking techniques have been introduced to accelerate inference time \cite{snerg, plenoxel, merf}. In pursuit of swifter training and inference, the utilization of diverse formats for learnable features yields more concise outcomes. While MLPs exhibit competitiveness in neural implicit representations \cite{inr_review} and original NeRFs, they lag behind methods that directly optimize and query the scene representations themselves in terms of speed. Notably, Instant-NGP \cite{ingp} employs multiresolution hash tables for scene representation organization, achieving fast processing with a level of detail. Other structures like  trees \cite{plenoctree}, and sparse grids \cite{plenoxel, DVGO} are also popular. In parallel, methods such as \cite{tensorf, ccnerf, nrff} structure the scene as tensor compositions with low rank. TensoRF \cite{tensorf}, for instance, employs vector-matrix (VM) and CANDECOMP/PARAFAC (CP) decomposition to characterize the scene, followed by input to compact MLPs for efficient computation. In a similar vein, CCNeRF \cite{ccnerf} discards the use of MLPs to facilitate compression and composition. NRFF extends TensoRF into a multiscale framework to enable comprehensive learning at varying scales, thereby enhancing accuracy.

\subsection{Neural Radiance Field with Multi-Date Images}
The vanilla NeRF model conditions on the fact that the 3D scene have the density, radiance and illumination, which is unreal for outdoor scenes, especially for internet photo collections and satellite images. Both of these two types of dataset have different lighting condition, appearance of the pixel at the same 3D position. Besides, there are transient occlusions from people and cars.
NeRF-W \cite{nerfw} firstly proposes a per-image latent embedding capturing photometric
appearance variations often present in wild heritage dataset, and explicitly model the scene into image-dependent and shared
components respectively,  to disentangle transient elements from the static scene. With the same starting point, Sat-NeRF adapt the method to satellite images by representing the albedo of the earth surface, and the other lighting factors in different heads of MLPs.
NeRF-OSR \cite{nerf-osr} enables outdoor scene relightning 

\subsection{Deep Learning in Reconstruction from Satellite Images}
Traditional 3D reconstruction has predominantly followed a multi-view stereo (MVS) pipeline that generates a dense disparity map using dense matching algorithms such as Semi-Global Matching (SGM) \cite{sgm}. However, errors tend to accumulate from the initial stereo pair selection to the subsequent refinement steps \cite{sgm, opt_sgm}. As deep learning methodologies infiltrate the realm of 3D reconstruction, an increasing number of end-to-end differentiable MVS pipelines have emerged \cite{mvsnet_review}. Some approaches extract features from the 3D cost volume and utilize 3D Convolutional Neural Networks (CNNs) \cite{mvsnet, casmvs}, Recurrent Neural Networks (RNNs) \cite{rnnmvs}, or attention modules \cite{attenmvs} to predict the disparity map. However, these methods often require significant computation and time. On the other hand, some methods adapt randomized iterative patch-match techniques \cite{gipuma, acmh} into an end-to-end framework \cite{pmnet, pmvsnet}.

The proliferation of Implicit Neural Representations (INR) introduces increased flexibility in terms of computation and memory utilization. Traditional epipolar geometry and INR complement each other in the realm of 3D reconstruction. MVS provides geometric guidance for the radiance field \cite{chen2021mvsnerf,nerfingmvs,dsnerf}, while INR reduces the reliance on conventional MVS pipelines for selecting optimal stereo pairs and explicit depth supervision \cite{srf, mvps}.

%

\section{Method}
\subsection{Light Field over the Earth Surface}
The primary objective of SatensoRF is to generate a synthesized novel view photo in RGB, along with an altitude map product.
The input of SatensoRF include the spatial coordinates of the scene volume, and the viewing direction $\mathbf{d}$.
For a point $\mathbf{x}_i=\mathbf{r}(t_i) = \mathbf{o}+t_i\mathbf{d}$, the proposed tensor radiance field yields both the rendered color $\mathbf{c}(\mathbf{r})$ and the rendered altitude $\mathbf{h}(\mathbf{r})$ for the ray $\mathbf{r}$ at a given sample $t_i$ within the interval $[t_{min}, t_{max}]$. Here, $t_{min}$ and $t_{max}$ define the sampling range along the altitude direction.

\begin{figure}
    \centering
    \includegraphics[width=8cm]{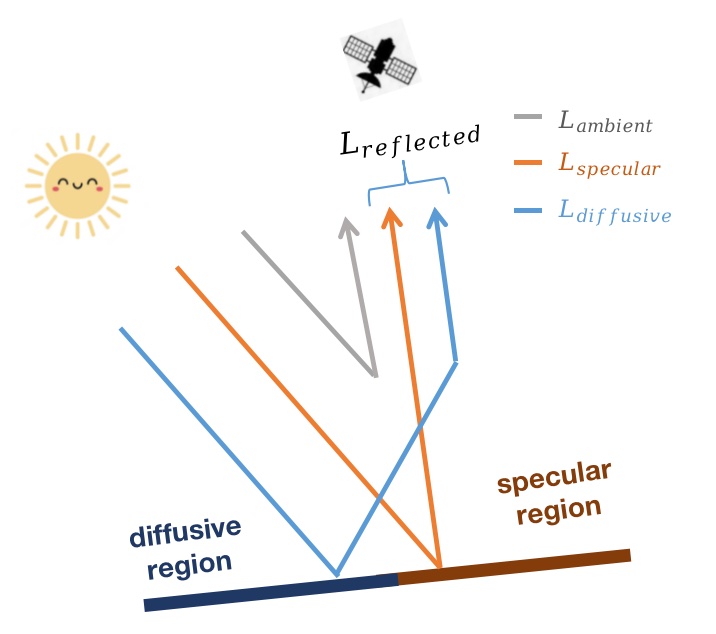}
    \caption{Sketch of the radiance components for the ground surface.}
    \label{fig:lightdecomp}
\end{figure}

The radiance observed by satellites over the atmosphere is the backscattered portion of incident solar radiation, which is composed of the pure backscattering and the radiance reflected by the underlying surface and transmittance to the satellite sensors.
The three principal components are illustrated in Fig.~\ref{fig:lightdecomp}:
\begin{enumerate}
    \item \textbf{ambient radiation} $L_{ambient}$: This pertains to photons that are scattered into the satellite's optical sensor without encountering the ground surface directly (depicted in gray).
    \item \textbf{specular radiation} $L_{specular}$: This denotes the specular components of the reflected radiation. 
    \item \textbf{diffusion radiation} $L_{diffuse}$:
    This involves the diffuse component of the reflected light.
\end{enumerate}
In this context, it's essential to emphasize that the ambient light component $L_{ambient}$ remains independent of the characteristics of the ground surface. Furthermore, the intricacies involved in tracing rays for scattered light $L_{diffuse}$ prompts us to omit the calculation of parameters for $L_{ambient}$ and $L_{diffuse}$ across every volume within the scene. Instead, we simplify these aspects through a unified irradiance model. Consequently, these factors blend in the overall radiation signal, which can be summarized as follows:
\begin{equation}
\begin{aligned}
    L_{all} 
    & = L_{ambient} + L_{diffuse} + L_{specular} \\ \notag
    &\triangleq L_{ambient} + L_{reflected} \\ \notag
\end{aligned}    
\end{equation}



 In contrast to S-NeRF and Sat-NeRF with heavy reliance on the solar direction, our approach resort to a dependent randomly initialized low-rank tensor field to decouple from the earth surface color. Instead, we rely on another small tensor field serving as the latent code randomly sampled from normal distribution, and train it to represent the ambient map.
 Ultimately, for the 3D point $\mathbf{x}_i$, we achieve the following composition:
\begin{align}
\mathbf{l}(\mathbf{x}_i) &= \lambda_{amb}(\mathbf{x}_i) \mathrm{l}_3 + (1-\lambda_{amb}(\mathbf{x}_i))
\mathbf{c}_{amb}(\mathbf{x}_i) \\ \notag
   \mathbf{c}(\mathbf{x}_i, \mathbf{d}) &= \mathbf{c}_{ref}(\mathbf{x}_i, \mathbf{d}) \odot  \mathbf{l}(\mathbf{x}_i)
\end{align}
where $\lambda_{amb}$ is the factor to adjust the impact of ambient radiation $L_{ambient}$ and the ambient radiation $L_{diffuse}$. The ambient color $\mathbf{c}_{amb}(\mathbf{x}_i)$ at $\mathbf{x}_i$ to form the irradiance map $\mathbf{l}(\mathbf{x}_i)$ to model $L_{diffuse}$, the reflected field $\mathbf{c}_{ref}$ to model $L_{reflected}$ will be discussed in detail in Section \ref{sec:tensorfield} .
So the final rendered map $\hat{I}$ is:
 \begin{align}
     \hat{I} &= \sum\limits_{i=1}^{N_s}Tr(\mathbf{x}_i)\alpha(\mathbf{x}_i)\mathbf{c}(\mathbf{x}_i, \mathbf{d})\\\notag&=\sum\limits_{i=1}^{N_s}Tr(\mathbf{x}_i)\alpha(\mathbf{x}_i)\mathbf{c}_{ref}(\mathbf{x}_i, \mathbf{d})\mathbf{l}(\mathbf{x}_i)
 \end{align}
 where the opacity $ \alpha(\mathbf{x}_i)$ and transmittance $Tr(\mathbf{x}_i)$ are calculated as 
 \begin{align}
     \alpha(\mathbf{x}_i) &= 1-\exp(-\sigma(\mathbf{x}_i)\delta_i\\
     Tr(\mathbf{x}_i) &= \prod\limits_{j=1}^{i-1}(1-\alpha(\mathbf{x}_j))
 \end{align} 
 with $\delta_i$ being the two consecutive points along the ray,  \ie $\delta_i = t_{i+1}-t_i$
 and $N_s$ is the number of sample in the altitude direction.

\subsection{Scene as Tensor Decomposition in LoD}
\label{sec:tensorfield}
\begin{figure}
\centering
\includegraphics[width=8cm]{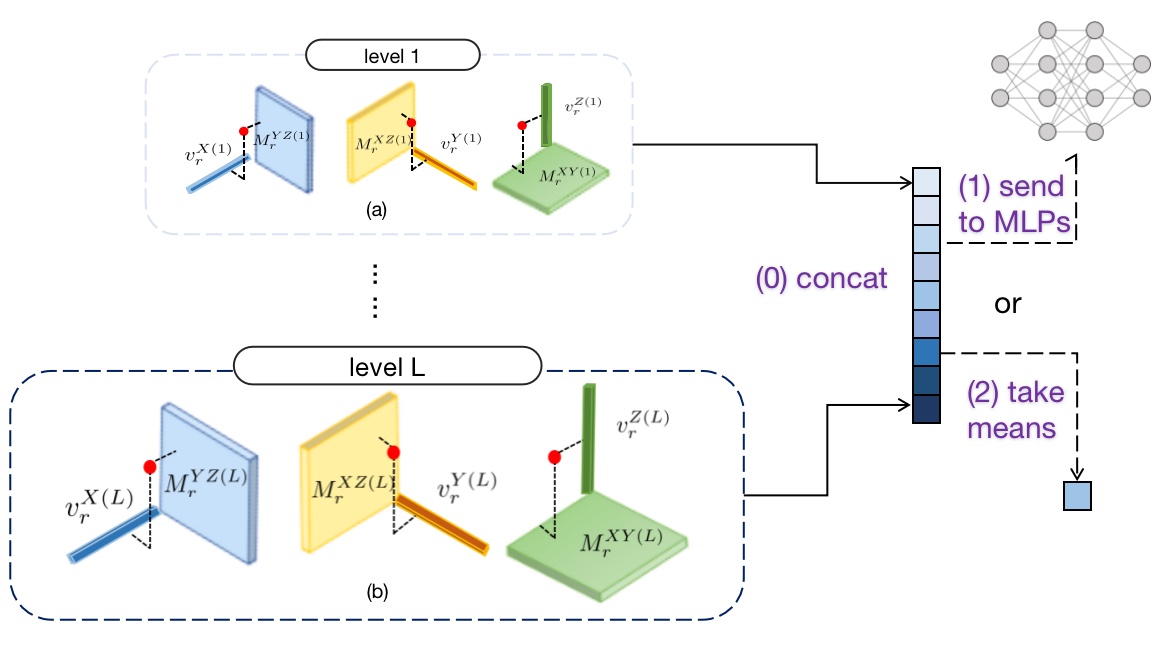}

\caption{Multiscale tensor factorization of the 3D scene before sending to the MLP}
\label{fig:VMinMS}
\end{figure}

Within our tensor radiance field, the atomic representation involves a singular level of VM decomposition, which factorizes a tensor into multiple vectors and matrices. This structure is illustrated in (a) of Fig.~\ref{fig:VMinMS}. We designate the three dimensions of the 3D space as $XYZ$, wherein $X$ and $Y$ correspond to the north and east directions in the Universal Transverse Mercator (UTM) coordinate system, while $Z$ signifies the elevation of the Earth's surface. Concretely, VM decomposition for a given single level $l$ can be formulated as follows:

\begin{equation}
   \mathcal{T}^{(l)}  = \sum\limits_{r=1}^{R}(v^{X(l)}_r \odot M_r^{YZ(l)} + v^{Y(l)}_r \odot M_r^{XZ(l)}+ v^{Z(l)}_r \odot M_r^{XY(l)})
\end{equation}
where $M^*_r$ are matrix factors for two
 of the three directions, $\odot$ is the element-wise multiplication and $R$ is the rank at three direction.

The coarse-to-fine strategy is a prevalent approach in NeRF pipelines. In our context, we have decomposed the feature field $r$ for the 3D scenes into $L$ levels. The sizes of each level are selected using a strategy akin to that of NGP \cite{ingp}, based on logarithmic considerations. Illustrations of the tensor factorization for level $1$ and level $L$ can be found in (a) and (b) of Fig.~\ref{fig:VMinMS}, respectively. It's important to note that distinct feature fields undergo varied procedures in subsequent stages.

\begin{figure*}s
\centering
\includegraphics[width=18cm]{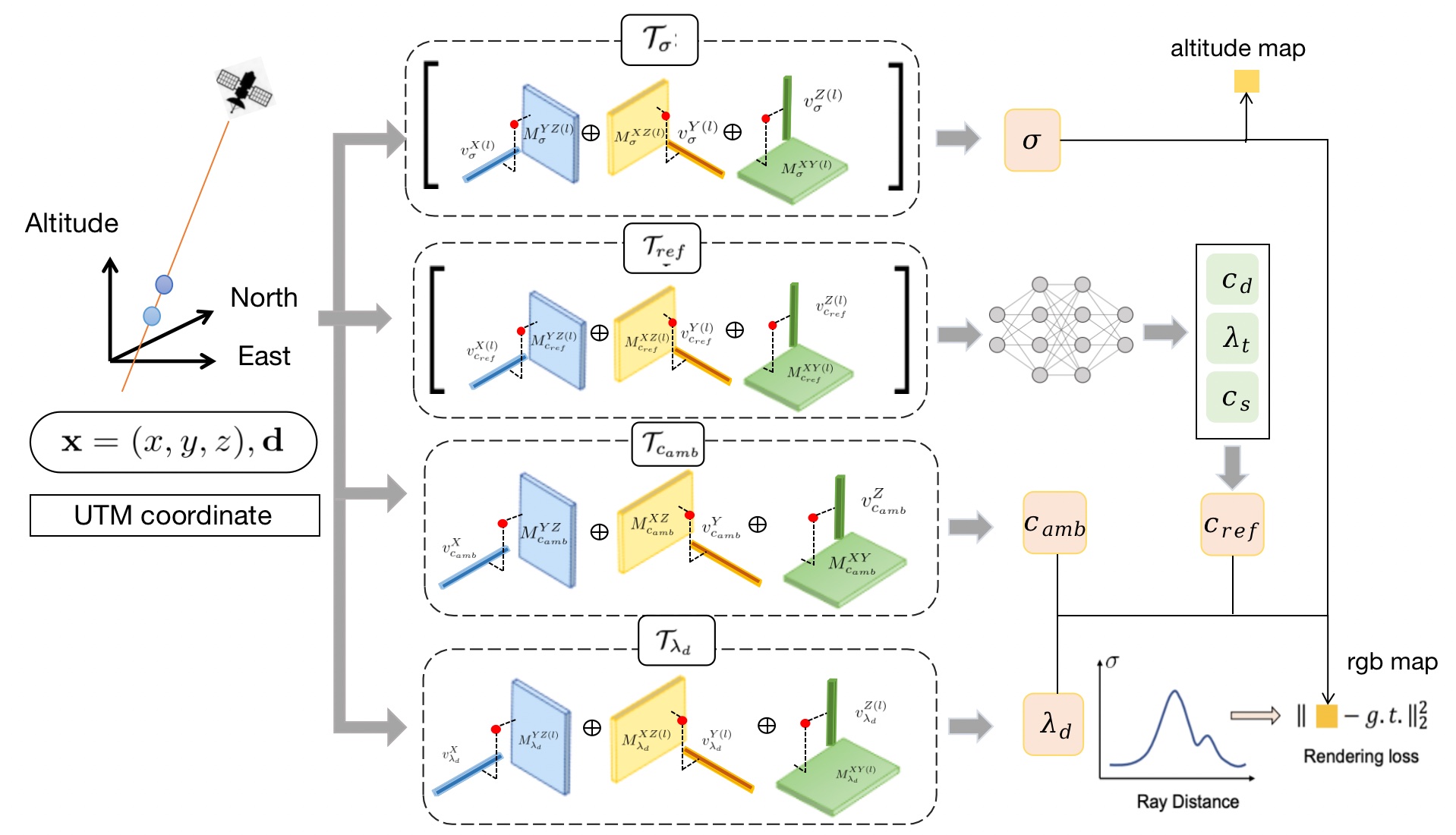}
\caption{The SatensoRF panorama encompasses four distinct sub-tensor radiance fields. These are outlined as follows:
1). The multi-scale volume density field $\mathcal{T}_{\sigma}$ is computed and then averaged to produce the volume density field.
2). The multiscale reflectance field $\mathcal{T}_{ref}$ is channeled into the MLPs for further processing.
3). Both the ambient color field $\mathcal{T}_{c_{amb}}$ and the ambient ambient adjusting factor $\mathcal{T}_{\lambda_{amb}}$ operate on a single-level scale. This is due to the fact that the finer details of the ground surface have minimal influence on ambient colors.
After volume rendering, the loss function is applied to the synthesized image to facilitate the optimization process. This panorama demonstrates the various components and their interactions within the SatensoRF framework.
}
\label{fig:pipeline}
\end{figure*}

To model both the radiance field and the irradiance field, SatensoRF employs feature fields that consist of several subfields, each with the following notations:
\begin{itemize}
    \item The volume density field $\mathcal{T}_{\sigma}$: This represents the radiance feature field for volume density. It has $L=8$ levels and $4$ feature channels. The outputs are concatenated and averaged across levels to form $\sigma$.
    \item The reflective field $\mathcal{T}_{ref}$: This represents the radiance feature field for appearance. It also has $L=8$ levels and $4$ feature channels. The outputs are concatenated and fed into a small multi-head MLP network, as shown in Fig.~\ref{fig:appfield}. Further details will be provided in Section \ref{sec:lightfield}.
    \item The ambient color field $\mathcal{T}_{c_{amb}}$: This represents the irradiance feature field for ambient color. It has only one level of tensor decomposition and consists of $3$ feature channels, without multiscale details. The outputs are concatenated and averaged across levels to form $\mathbf{c}_{amb}$.
    \item The ambient factor field $\mathcal{T}_{\lambda_{amb}}$: This represents the irradiance field feature field. It has only one level of tensor decomposition and includes $2$ feature channels, without multiscale details. The outputs are concatenated and averaged across levels to form $\lambda_{amb}$.
\end{itemize}

\subsection{The Reflective Radiance Field}
\label{sec:lightfield}
The result color $\mathbf{c}_{ref}$ of the reflective light $L_{reflected}$ can be decomposed into diffusion color $\mathbf{c}_d$ and specular light $\mathbf{c}_s$ with factor $\lambda_s$ \cite{rendering_eq,pbrbook}:
\begin{align}
    \mathbf{c}_{ref}(\omega_o;\mathbf{x}) & = \mathbf{c}_{d} + \lambda_s \mathbf{c}_{s} \\ \notag
    & = \mathbf{c}_{d} + \lambda_s \int_{\omega} L_i(\omega_i;\mathbf{x})\rho_s(\omega_i,\omega_o;\mathbf{x})(\mathbf{n}\cdot \omega_i)d\omega_i \\ \notag
    & = \mathbf{c}_{d} + \lambda_s \int_{\omega}f(\omega_i,\omega_o;\mathbf{x},\mathbf{n})d\omega_i
    \label{eq:brdf}
\end{align}
where $L_i$ indicates the incoming radiance from direction $\omega_i$ around the surface normal $\mathbf{n}$ at position $\mathbf{x}$, $\rho_s$ is the bidirectional reflectance distribution function (BRDF) depending on $\mathbf{x}$ and $\omega_i$, and the dot product between $\omega_i$ and $\mathbf{n}$ serves as the cosine of the intersecting angle. The specular part is the integration over the hemisphere $\Omega$ centered around $\mathbf{n}$ containing all possible values for $\omega_i$ where $\omega_i \cdot \mathbf{n} > 0$. Instead of solving the rendering equation via Monte Carlo estimation, implicit representations learn the components of the rendering equation  via neural networks \cite{tensoir,nrff}.

\begin{figure*}
\centering
\includegraphics[width=15cm]{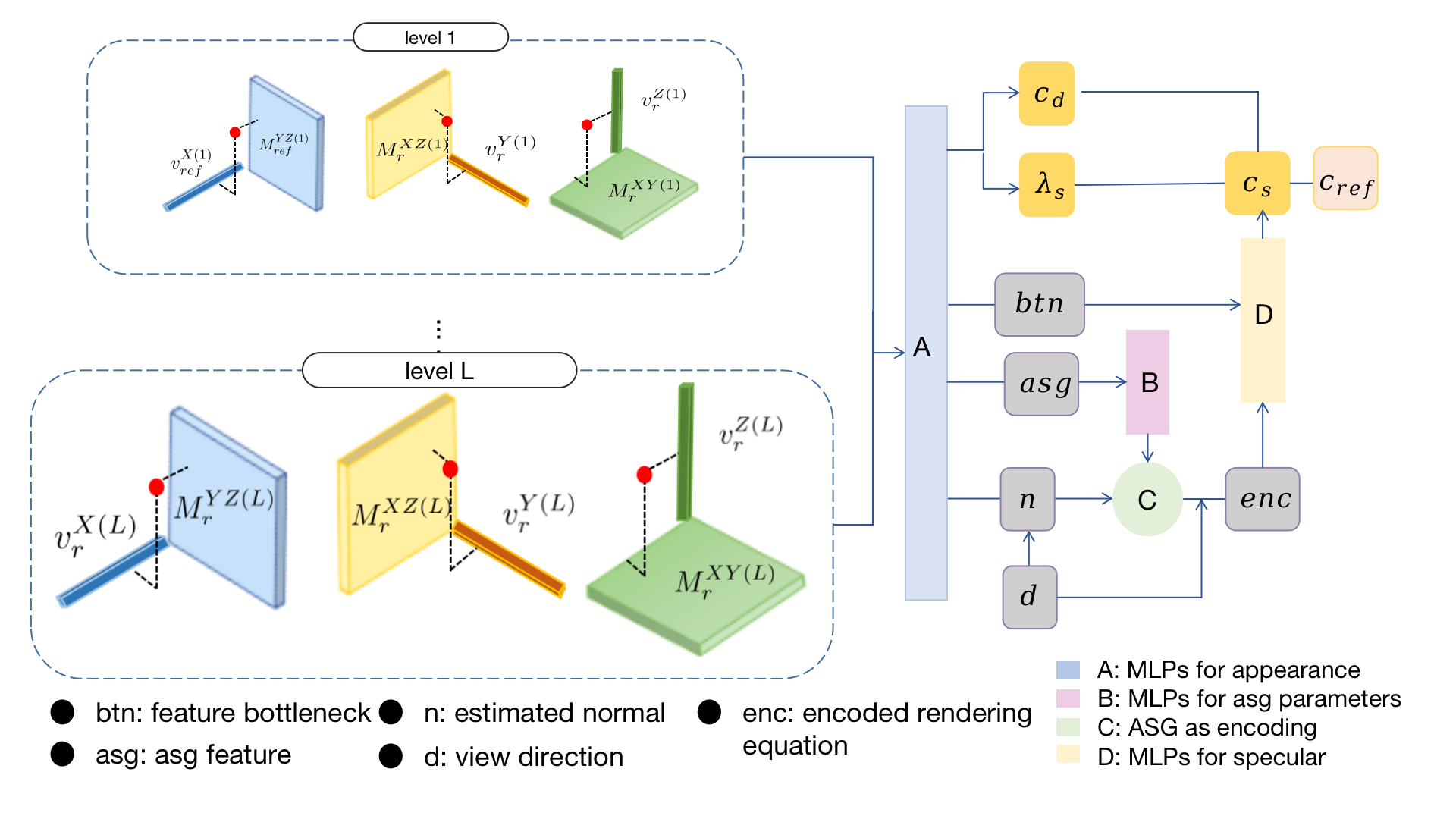}
\caption{The multihead MLPs to decouple the reflective light field.}
\label{fig:appfield}
\end{figure*}

Subsequently, we perform a decomposition of the reflective radiance field using the ambient color feature field $\mathcal{T}_{c_{amb}}$. We then feed the extracted features into Multi-Layer Perceptrons (MLPs). According to Equation~\ref{eq:brdf}, both the diffusion color $c_d$ and the factor $\lambda_s$ remain independent of the viewing direction. As a result, they can be directly regressed through MLPs. In this case, a 4-layer MLP structure (labeled as "A" in Fig.~\ref{fig:appfield}) is employed to predict the diffusion color $c_d$, the factor $\lambda_s$, and the normals $n$ for every volume intersected by the sampled ray.

To improve the parameterization of the rendering equation, the encoding process is applied within the color feature space, which differs from most NeRF methods that perform encoding in the scene coordinate space \cite{nerfreview, nrff}. This approach is particularly advantageous because the BRDF function $f$ depends solely on the specific features of the color domain.
The encoded feature of the rendering equation is modeled as an Anisotropic Spherical Gaussian (ASG) function, a concept introduced by the work \cite{asg}. This ASG function is composed of subfunctions $G_i$ that further enhance the ability to represent intricate scene characteristics:
\begin{align}
    &\mathcal{F}_s(\omega_o, \mathbf{x}) = \sum\limits_{i=0}^{N-1}G_i(\omega_o;\mathbf{x}, [\omega_i,\omega_i^{\lambda}, \omega_i^{\mu}], [\lambda_i,\mu_i], a_i)\\ \notag
    & = \sum\limits_{i=0}^{N-1} F_i \max(\omega_o\cdot\omega_i,0)\exp(-\lambda_i(\omega_o\cdot\omega_i^{\lambda})^2 - \mu_i(\omega_o\cdot\omega_i^{\mu})^2)
\end{align}
where $F_i$ is the feature vector, $\mathcal{B} = [\omega_i,\omega_i^{\lambda}, \omega_i,\omega_i^{\mu}]$ are orthonormal basis with $\lambda_i, \mu_i>0$ being the bandwidth at each basis.

The parameters of the ASG $F_i$, $\lambda_i$, and $\mu_i$, are initially obtained from two MLPs. The first MLP labeled "A" (blue) and another 2-layer MLP labeled "B" (pink) in Fig.~\ref{fig:appfield} contribute to determining these ASG parameters.
The basis $\mathcal{B}$ required for ASG representations is sampled on a unit sphere. 
These basis vectors are then processed through the encoder labeled "C" (green) in Fig.~\ref{fig:appfield}. The encoder generates the rendering equation that is contingent on the viewing direction, together with the intersection angle derived from the dot product of the normals $n$ and the viewing direction $d$.
Subsequently, the specular part $\mathbf{c}_s$ is decoded using the last 2-layer MLP labeled "D" in Fig.~\ref{fig:appfield}, employing a Sigmoid activation function. This decoded specular part $\mathbf{c}s$ then contributes to forming the reflective color $\mathbf{c}_{ref}$. 


\subsection{Handling Transient Objects in Density Reconstruction}
\label{sec:total_variation}
Instead of modeling the effects caused by transient objects by regressing the uncertainty, as proposed in \cite{satnerf} \cite{satnerf}, we capitalize on the inherent advantages of low-rank tensor representations.

Total Variation Denoising (TVD) is a widely employed method for signal restoration tasks. This regularization technique is particularly effective in preserving sharp edges present within the underlying signal.

In this context, we treat the Earth's ground surface as the signal to be restored from the noisy input data. Meanwhile, transient objects such as cars or people are considered as the noise to be filtered out.

We denote the true volume density tensor as $\mathcal{T}_{\sigma}$ and subsequently formulate the volume density tensor as follows:

\begin{equation}
    \mathcal{T}_{\sigma} = \mathcal{X}_{\sigma} + \mathcal{N}
    \label{eq:sigma_decomp}
\end{equation}
 where $\mathcal{N}$ is the noise tensor caused by the transient object with no prior rules of occurrence.

Remote sensing images often exhibit a prominent piecewise smooth structure, making them well-suited for the application of Total Variation (TV) methods in image processing \cite{tvdenoise2015, tvdenoise2019}. The underlying rationale of Total Variation regularization is rooted in the notion that excessive and spurious details tend to exhibit higher total variation \cite{lowrank_review, brief_im_denoise}.

In light of this, we apply the concept of Total Variation loss to the decomposited planes of the tensor field $\mathcal{T}_{\sigma}$ across all $L$ levels. This approach is geared towards promoting smoother and more coherent structures within the data.
\begin{align}
 \mathcal{L}_{TV}(\mathcal{T}_{\sigma})   & = \sum\limits_{l=1}^L \|\mathbf{D} \mathcal{T}_{\sigma}\|_2 / L\\
 \|\mathbf{D}\mathcal{T}_{\sigma}^{(l)}\|_2&=\sum\limits_{l=1}^L(\sum\limits_{i}(M_{\sigma}^{XY(l)}(i+1,:) - M_{\sigma}^{XY(l)}(i,:)) \\ \notag
 & + \sum\limits_{j}(M_{\sigma}^{XY(l)}(:,j+1) - M_{\sigma}^{XY(l)}(:,j)))
\end{align}
where $\mathbf{D}$ is the difference operator, and $i, j$ indexes elements on the factorized matrix $M_{\sigma}^{XY(l)}$ at scale $l$ from the $L$ levels. 

\subsection{The Losses}
\paragraph{$\mathcal{L}_{rgb}$}
The basic supervisions is the mean squared error
(MSE) between the rendered color at every ray $\mathbf{R}$ and the ground truth color of the input images:
\begin{equation}
    \mathcal{L}_{rgb} = \sum\limits_{\mathbf{r}in\mathcal{R}}\|\hat{\mathbf{c}}(\mathbf{r}) - c_{gt}((\mathbf{r})\|_2^2
\end{equation}
\paragraph{$\mathcal{L}_{normal}$}
Additionally we should force the estimated normals ${\mathbf{n}_i}$ to be on the unit sphere:
\begin{equation} 
\mathcal{L}_{normal} = \frac{1}{N}\sum\limits_{i=0}^{N-1} \omega_i \max(0, \mathbf{d}\cdot\mathbf{n}_i)^2 
\end{equation}
where $N$ is the number of samples in the cast ray and $\mathbf{d}$ is the viewing direction.
\paragraph{$\mathcal{L}_{\lambda_{amb}}$}
The ambient light $L_{ambient}$ takes no effect when the ray intersects with the earth surface, thus we encourage the  ratio of the ambient light $\lambda_{amb}$ to be close to the transmittance $Tr$. 
\begin{equation}
   \mathcal{L}_{\lambda_{amb}} = \sum\limits_{i=1}^N(Tr(x_i)-\lambda_{amb}(x_i))^2 + 1-\sum\limits_{i=1}^NTr(x_i)\alpha(x_i)\lambda_{amb}(x_i)
\end{equation}
where $N$ is number of total voxels.
\paragraph{$\mathcal{L}_{depth}$}
To further enhance model performance, particularly concerning altitude reconstruction, SatensoRF incorporates depth supervision, akin to Sat-NeRF and DS-NeRF \cite{dsnerf}. Utilizing precomputed sparse 3D points derived from SFM pipelines or bundle-adjusting procedures, the depth supervision is formulated as follows:
\begin{equation}
    \mathcal{L}_{DS} = \sum\limits_{\mathbf{r}\in \mathcal{R}_{DS}} w(\mathbf{r})(\hat{H}(\mathbf{r})-\|\mathbf{X}(\mathbf{r})-\mathbf{o}(\mathbf{r})\|_2)^2
    \label{eq:ds}
\end{equation}
where $\mathcal{R}_{DS}$ is the set of rays containing the known 3D points, $\|\mathbf{X}(\mathbf{r})-\mathbf{o}\mathbf{r}\|_2$ is the relative altitude as ground truth at ray $\mathbf{r}$, $\hat{H}(\mathbf{r})$ is the rendered height as estimation, and $w(\mathbf{r})$ comes from the reprojection error from the bundle adjustment procedure \footnote{https://github.com/centreborelli/sat-bundleadjust}.

\paragraph{Total loss}
 The basic whole loss include the data fidelity term $\mathcal{L}_{rgb}$, the noise sparse prior $\mathcal{L}_{TV}$, the normal regularization $\mathcal{L}_{normal}$ and the ambient light regularization $\mathcal{L}_{\lambda_{amb}}$:

\begin{equation}\label{eq:total_loss}
\begin{aligned}
    \mathcal{L} =& \mathcal{L}_{rgb}(\mathbf{c}, c_{gt}) + \lambda_{TV}\mathcal{L}_{TV}(\mathcal{T}_{\sigma})\\ \notag
    & + \lambda_{n}\mathcal{L}_{normal} + \lambda_{\lambda_{amb}}\mathcal{L}_{\lambda_{amb}}   
\end{aligned} 
\end{equation}

\section{Experiments}
We evaluate the performance of SatensoRF based on the subsets of spacenet MVS-dataset \footnote{https://spacenet.ai/iarpa-multi-view-stereo-3d-mapping/}. 

The evaluation is conducted from the following aspects. 
\begin{enumerate}  
    \item For both single-date and multi-date images, we validate the effectiveness of SatensoRF, and compare it to the state-of-the-art Sat-NeRF and other stereo pipelines in Section \ref{sec:comp_performance} and Section \ref{sec:comp_efficiency}.
    
    \item We validate the non-Lambertian assumption for SatensoRF in Section \ref{sec:validate_nonlamber}.

    \item We conduct ablation study on total variation loss for SatensoRF in Section \ref{sec:validate_tv}.
    
    \item We conduct controled experiments to explore the effects of changing some variables in Section \ref{sec:controlled_exp}.
    
\end{enumerate}

\subsection{Setup}
\subsubsection{Dataset Preparation}
Firstly for multi-date images, we try to validate that SatensoRF applies well to the large images with minimal available primitive information.
So we choose the original dataset from the Maxar WorldView-3 satellite, capturing imagery between 2014 and 2016 over the city of Jacksonville, Florida, in the United States. In contrast to the multi-date images used by S-NeRF and Sat-NeRF, our experiments are conducted using full-size RGB images, each containing over 4 million pixels. This approach differs from cropping images into smaller patches based on areas of interest (AOI).
 
The original dataset \footnote{https://spacenet.ai/iarpa-multi-view-stereo-3d-mapping/} comprises multi-date imagery captured under various lighting conditions, and specific details can be found in Table~\ref{tab:dataset}. The original multi-date dataset contains approximately ten times as many rays as the DFC dataset, and it includes challenging scenarios. For instance, Subset 031 features a mix of vegetation, buildings, and asphalt roads, Subset 159 includes a reflective flyover, and Subsets 260 and 280 contain bodies of water. 

Notably, Not all of the subsets of Jacksonville has corresponding multispectral images (MSI) to offer the extra information about sun elevation and sun azimuth, so we pick up two subsets with extra sun information (017 and 159), and two of subsets (260 and 280) without other clues. 
Specificaly, the extra information of solar elevation and solar azimuth serve as the input of sun position in the compared sat-nerf models. For subset 017 and 159 we take these extra sun information as sat-nerf input, and for subset 260 and 280 we take $45^\circ$ as azimuth, and $90^\circ$ elevation angle as the sun position placeholder for sat-nerf model. 

Actually, for the sake of fairness, we shall keep all the input the same, like the process for subset 260 and 280. While here we make some concessions on subset 031 and 159 to validate the capability of SatensoRF for handling unknown sun position against Sat-NeRF with known sun positions.

\begin{figure}
    \centering
    \includegraphics[width=8cm]{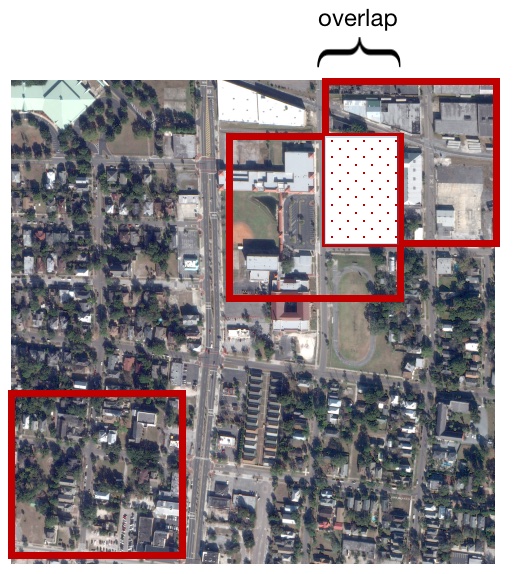}
    \caption{Illustration of cropping single-date imagery with fixed-size of overlapping to support rational scene geometry for nerf-based method training.}
    \label{fig:crop_single_date}
\end{figure}

\begin{table*}[t]
    \centering
        \caption{The dataset comparisons. On the MVS-dataset from spacenet, we create two types of datasets to study how SatensoRF acts on single-date imagery and multi-date imagery. \# indicates the count of number, and the measure of overlap is pixel. For single-date dataset, we crop the RGB image of index 006 into $512\times 512$ patches, and for multi-date dataset, we use the images of the original sizes. \checkmark indicates the existence of the elevation and azimuth angle of sun position. Alt. bounds indicates the altitude range of the area within the image.}
    \begin{tabular}{c|cccccccccc}
    \toprule
   & subset & Resolution & \#Rays & \#Images & overlap[pix]&\#train&\#test&elevation&azimuth&Alt. bounds [m]  \\
    \midrule
        \multirow{4}{*}{\thead{single\\-date}}
      &  017-6&$512\times512$&$262604$ & 41 &  0.7813&18 &2 &\checkmark& \checkmark& [-23, 4] \\
      &033-6 &$512\times512$ &$262604$ &41 & 0.6836&18&2&\checkmark& \checkmark& [-27, 3] \\
      &072-6  &$512\times512$ &$262604$ &41 & 0.5859&18&2&\checkmark& \checkmark& [-22, 1] \\
      &251-6  &$512\times512$ & $262604$&41 & 0.4883&18&2&\XSolidBrush&\XSolidBrush& [-28, 14] \\
      
        \hline
         \multirow{4}{*}{\thead{multi\\-date}}
         &017&$2400\times 2400$&$5760000$ &10& - &9&1&\checkmark& \checkmark& [-23, 4]\\
         &159&$2400\times 2400$& $5760000$& 11 & - &10&1&\checkmark& \checkmark& [-29, 15]\\
         &260&$2400\times 2400$& $5760000$& 16& - &14&2 &\XSolidBrush&\XSolidBrush& [-30, 13]\\
         &280& $2400\times 2400$& $5760000$ &24& - & 21&3 &\XSolidBrush&\XSolidBrush& [-29, 3]\\
         \hline
         \multirow{4}{*}{\thead{DFC}}
         &004& $780\times 765$ & $596276$ &11 & -&9&2&\checkmark& \checkmark & [-24, 1]\\
         &068&$795\times 771$ & $613480$ &19 & - &17&2&\checkmark& \checkmark& [-27, 30]\\
         &214& $790\times 761$ &$601292$ &24& - &21&3&\checkmark& \checkmark& [-29, 73]\\
         &260&$794\times 771$ & $612412$& 17& - &15&2&\checkmark& \checkmark& [-30, 13]\\
    \bottomrule
    \end{tabular}

    \label{tab:dataset}
\end{table*}
Moreover, we also crop each subsets into overlapped multi-view imagery of size $512\times512$, to evaluate model performance on the same date. For every chosen original subsets of spacenet dataset, we crop the image indexed by 006 in the subset, namely 017-6, 033-6, 072-6 and 251-6.  The cropping mode is shown in Fig.~\ref{fig:crop_single_date}. And for single-date imagery we also use a subset 251-6 without extra sun information.

\subsubsection{Implemented Detail}
All the SatensoRF and its variants are implemented in PyTorch. For the baseline model we set $L=16$ levels of tensor decomposition for both the appearance field $\mathcal{T}_{ref}$ and the density field $\mathcal{T}_{\sigma}$,
$L=1$ level for the sky color field $\mathcal{T}_{\sigma}$. The coefficient for the RGB loss $\mathcal{L}_{rgb}$ and the total variation loss  $\mathcal{L}_{TV}$ are both $1.0$. All the learnable parameters are optimized with Adam optimizer with a learning rate of $0.0002$ for the tensor structure and $0.0001$ for the MLPs. Both of the learning rates degrade log-linearly to 0.1
times to their values in the latest steps. The batchsize is set to 8192, and all the experiments are conducted on a single card of  NVIDIA GeForce RTX 3090 GPU with 16 GB RAM.


\subsection{Efficiency Comparison}
\label{sec:comp_efficiency}
Firstly, we show that SatensoRF precedes other satellite neural radiance field in efficiency in both training and inference phase. For multi-date dataset, we stop training when the validation PSNR starts to fall, and all of the time are taken average over different subsets of dataset. While for single-date dataset, since PSNR increases fast and we stop training after the validation PSNR reaches 36. Here we basically take the measures below to show the computation efficiency:
\begin{enumerate}
    \item Training time $T_{train}^{multi}(s)$ on multi-date dataset in seconds, until the validation PSNR starts to fall.
        \item Training time $T_{train}^{single}(s)$ on multi-date dataset in seconds, until the validation PSNR reaches $36dB$.
    \item Inference time $T_{test}^{multi}(s)$ of computing a total image from the multi-date dataset in seconds.
        \item Inference time $T_{test}^{single}(s)$ of computing a total image from the single-date dataset in seconds.
    \item Maximal setting of batchsize $BS_{max}$  for the model to reach the maximum of  the single-card GPU memory.
    \item Number of parameters $\#Params$: number of the total parameters with $M$ indicating a million.
    \item $FLOPs^{multi}$ (floating point operations) for multi-date imagery with $B$ indicating a billion.  
    \item $FLOPs^{single}$ (floating point operations) for single-date imagery with $B$ indicating a billion.  
\end{enumerate}
The basic comparisons are listed in Table~\ref{tab:efficiency}. As can be seen, the full-size SatensoRF model has roughly a quater of parameters of Sat-NeRF models, with the memory cost are far more less during training and testing, so the batchsize of SatensoRF can be set to nearly 8 to 16 times of the Sat-NeRF batchsize. Besides, the computation of SatensoRF are roughly one third of Sat-NeRF measured in FLOPs.


\begin{table*}
\centering
    \caption{Efficiency Comparison of SatensoRF and the state-of-the-art methods. $M$ is the abbreviation for million, and $B$ is the abbreviation for billion.}
    \begin{tabular}{c|cccccccc}
    \toprule     &$T_{train}^{multi}(s)$&$T_{train}^{single}(s)$&$T_{test}^{multi}(s)$ &$T_{test}^{single}(s)$& $BS_{max}$ & $\#Params(M)$&$FLOPs^{multi}(B)$&$FLOPs^{single}(B)$\\
         \midrule
        NeRF & 26648.04&14481.13&131.47& 17.23&1024&2.30&26462.45&1204.34\\
        \hline
        S-NeRF &28021.71&12280.29&136.76& 18.02&1024&2.50&28768.67&1309.30 \\
        \hline
        Sat-NeRF & 37129.39&13826.52&141.27&19.31&1024&2.64&30293.36&1378.68  \\      
        \hline
        SatensoRF &\textbf{10129.81} &\textbf{2069.65}& \textbf{97.49}&\textbf{9.35}& \textbf{8096}&\textbf{0.79}&\textbf{416.21}&\textbf{416.21} \\

    \bottomrule
    \end{tabular}

    \label{tab:efficiency}
\end{table*}
\begin{table*}
    \centering
        \caption{Numerical results comparisons of SatensoRF and the SOTA work on multi-date images. Postfix \textit{-SC} refers to models with solar correction supervision in Sat-NeRF paper \cite{satnerf}, and \textit{-DS} refers to models with depth supervision. }
    \begin{tabular}{c|cccc|cccc|cccc}  
    \toprule 
    &
    \multicolumn{4}{c}{\textbf{Test PSNR}(dB)$\uparrow$}
    &\multicolumn{4}{c}{\textbf{Test SSIM}$\uparrow$}
    &\multicolumn{4}{c}{\textbf{Test Altitude MAE}(m)$\downarrow$}    
    \cr  
    \cmidrule{2-13}
    Area Name&031&159&260&280&031&159&260&280&031&159&260&280\cr 
    \midrule
     NeRF& 19.595 &18.794 &16.876& 18.921& 0.716 &0.761& 0.483 &0.703& 3.449& 2.628&-&-\cr 
    \midrule[1pt]
    S-NeRF + SC &21.059& 22.013&17.036 &20.514 & 0.787& 0.798&0.504&0.711 & 2.848&2.318&-&-\cr 
    \midrule[1pt]
    Sat-NeRF + SC &21.090 & 22.397& 20.710 &20.989 &0.792 &0.891& 0.770&0.783 & 2.722 &1.861&-&-\cr 
 \hline
Sat-NeRF + DS + SC& 21.617& 23.812& 21.094 &21.322& 0.801 &0.901 &0.792&0.813 & \textbf{1.982} &\textbf{1.351}&-&-\cr 
\midrule[1pt]
SatensoRF + SC& 24.629&26.695&22.706 & 24.578&0.921&0.933&0.918 &0.939 & 2.943& 2.081&-&-\cr 
\hline
SatensoRF + DS + SC& \textbf{24.771}&\textbf{27.013}&\textbf{23.151}&\textbf{24.866}&\textbf{0.934}&\textbf{0.929}&\textbf{0.952}&\textbf{0.931}&1.998&1.859&-&-\cr 

    \bottomrule  
    \end{tabular}   

    \label{tab:comp2sota_multi-date}
\end{table*}
\begin{table*}
    \centering
    \caption{Numerical results comparisons of SatensoRF and the SOTA works on the seen cases of multi-date imagery. Postfix \textit{-SC} refers to models with solar correction supervision in Sat-NeRF paper \cite{satnerf}, and \textit{-DS} refers to models with depth supervision.}
    \label{tab:comp2sota_seen}
    \begin{tabular}{c|cccc|cccc|cccc}  
    \toprule   
    &
    \multicolumn{4}{c}{\textbf{Train PSNR}(dB)$\uparrow$}
    &\multicolumn{4}{c}{\textbf{Train SSIM}$\uparrow$}
    &\multicolumn{4}{c}{\textbf{Train Altitude MAE}$(m)\downarrow$}    
    \cr  
    \cmidrule{2-13}
    Area Name&031&159&260&280&031&159&260&280&031&159&260&280\cr 
    \midrule
     NeRF& 26.342 &23.847 &27.601& 24.905& 0.723 &0.805&0.753 &0.823& 3.357& 2.051&-&-\cr 
   \midrule[1pt]
    S-NeRF + SC &22.901&26.088 & 18.914& 21.018& 0.799& 0.802&0.646&0.724 & 2.314&2.009&-&-\cr 
    \midrule[1pt]
    Sat-NeRF + SC &23.909&24.105&23.782&21.972&0.835&0.914&0.881&0.831&2.531&1.779&-&-\cr 
 \hline
Sat-NeRF + DS + SC& 24.101&24.597&24.710&22.849&0.821&0.909&0.804&0.821&\textbf{1.759}&\textbf{1.104}&-&-\cr 
\midrule[1pt]
SatensoRF + SC& 29.639&32.705& 30.988& 29.618&0.932&\textbf{0.958}& 0.924&\textbf{0.951} & 2.815& 2.019&-&-\cr 
\hline
SatensoRF + DS + SC& \textbf{29.872}&\textbf{33.382}&\textbf{31.210}&\textbf{29.905}&\textbf{0.956}&0.947& \textbf{0.962}&0.949 &1.762& 1.596&-&-\cr 

    \bottomrule  
    \end{tabular}   

\end{table*}

\begin{figure*}
      \centering
    \includegraphics[width=18cm]{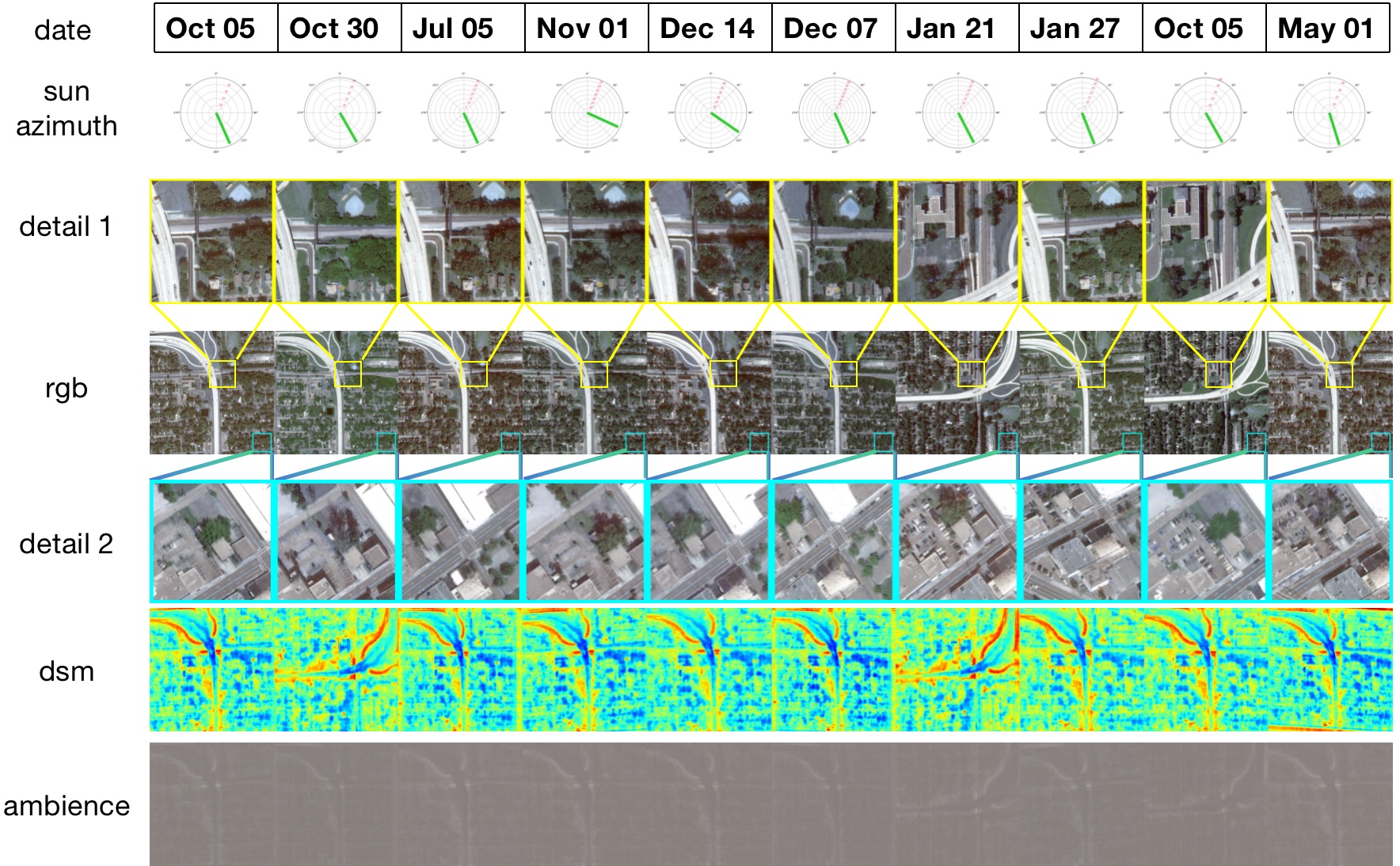}
    \caption{Details of the synthesized images in subset 031, with details of street views of size $500\times 500$ and  details of  vegetations of size $300\times 300$.}
    \label{fig:031}  
\end{figure*}
\subsection{Performance Comparison}
\label{sec:comp_performance}
The output of SatensoRF is twofold, a synthesized novel view and an altitude map, so we compare SatensoRF to the extant neural radiance fields for satellite with satellite cameras.
The comparisons mainly include vanilla NeRF \cite{nerfreview}, S-NeRF \cite{snerf}, Sat-NeRF \cite{satnerf}. Since the experiments in \cite{satnerf} has validated the effectiveness of solar correction loss for Sat-NeRF, we only take Sat-NeRF with solar-correction (SC) for comparison to yield better results, and the notations are clarified  below:
\begin{itemize}
    \item S-NeRF + SC: S-NeRF with solar correction.
    \item Sat-NeRF + SC: Sat-NeRF with solar correction.
    \item Sat-NeRF + DS: Sat-NeRF with depth supervision.
    \item Sat-NeRF + DS + SC: Sat-NeRF with solar correction and depth supervision.
         \item SatensoRF + SC: SatensoRF with Solar Correction.
\end{itemize}
\subsubsection{Multi-date Comparison}
Table \ref{tab:comp2sota_multi-date} presents the multi-date comparison of SatensoRF and the SOTA neural radiance field works for multi-date imagery on test images unseen by the trained model. With shorter training time and inference time, SatensoRF achieves better results in novel view synthesis task with PSNR and SSIM as metrics, and the quality of the reconstructed DSM products equals to Sat-NeRF by and large. The MAE of subset 260 and 280 are ommited because there are water within the area, and we did not perform exact segmentations to find the accurate areas to compute DSM.

While Table \ref{tab:comp2sota_seen} presents the multi-date comparison for multi-date imagery on the training images. As shown in Table \ref{tab:comp2sota_seen}, SatensoRF converges faster to higher PSNR than all of the NeRF-based models, and the SSIM has increased more during shorter training time, synchronously with PSNR.
The synthesized results with details of subsets 031, 159 are shown in Fig.~\ref{fig:031}, Fig.~\ref{fig:031} respectively. The first and second rows show the different dates and the sun azimuth of the first 10 images synthesized by SatensoRF. The $3rd$ and the $5th$ row shows two levels of details in the synthesized images. The $6th$ row shows the estimated altitude map from the synthesized DSM, while the last row shows the ambient map ($c_{amb}$) of different date. 

Both of the two tables come to two main conclusion.
Firstly, for multi-date images of original large size, SatensoRF has better performance than Sat-NeRF and previous methods.
Secondly, from the results on subset 260 and 280, even though the input for SatensoRF has no solar information, the novel view synthesizing tasks still takes the lead in PSNR/SSIM.

\begin{figure*}
          \centering
    \includegraphics[width=18cm]{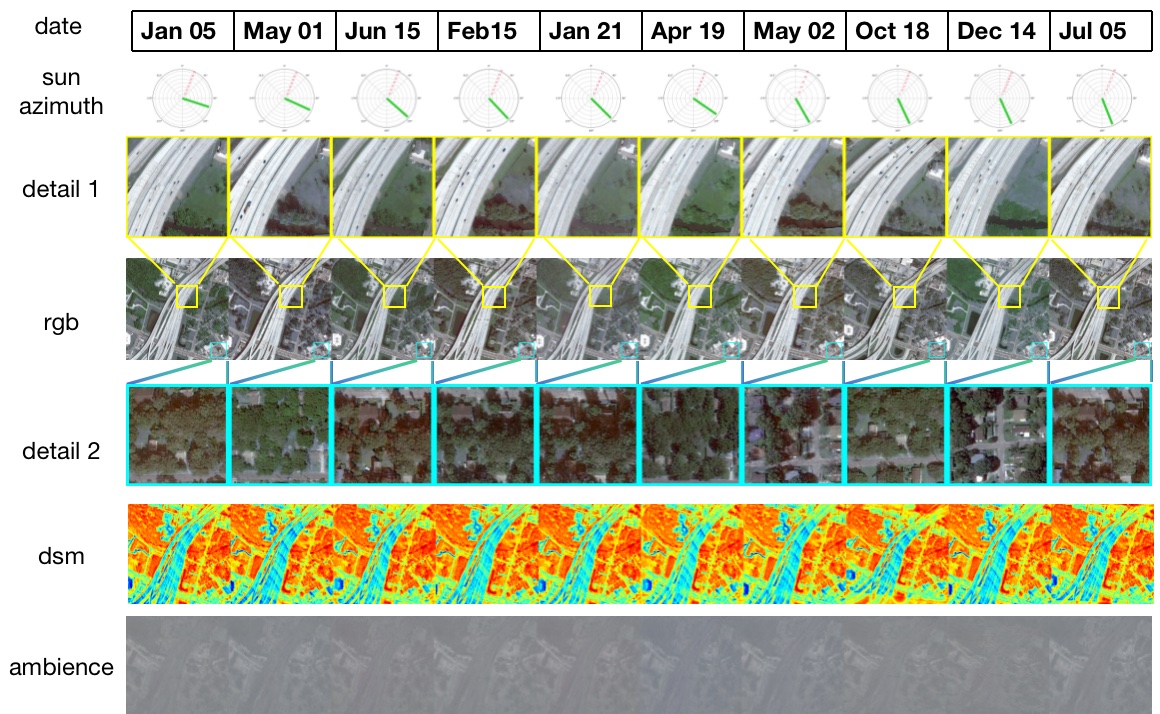}
    \caption{Details of the synthesized images in subset 159, with details of asphalt flyover of size $500\times 500$ and  details of  street views of size $300\times 300$.}
    \label{fig:159}
\end{figure*}
\begin{figure*}
          \centering
    \includegraphics[width=18cm]{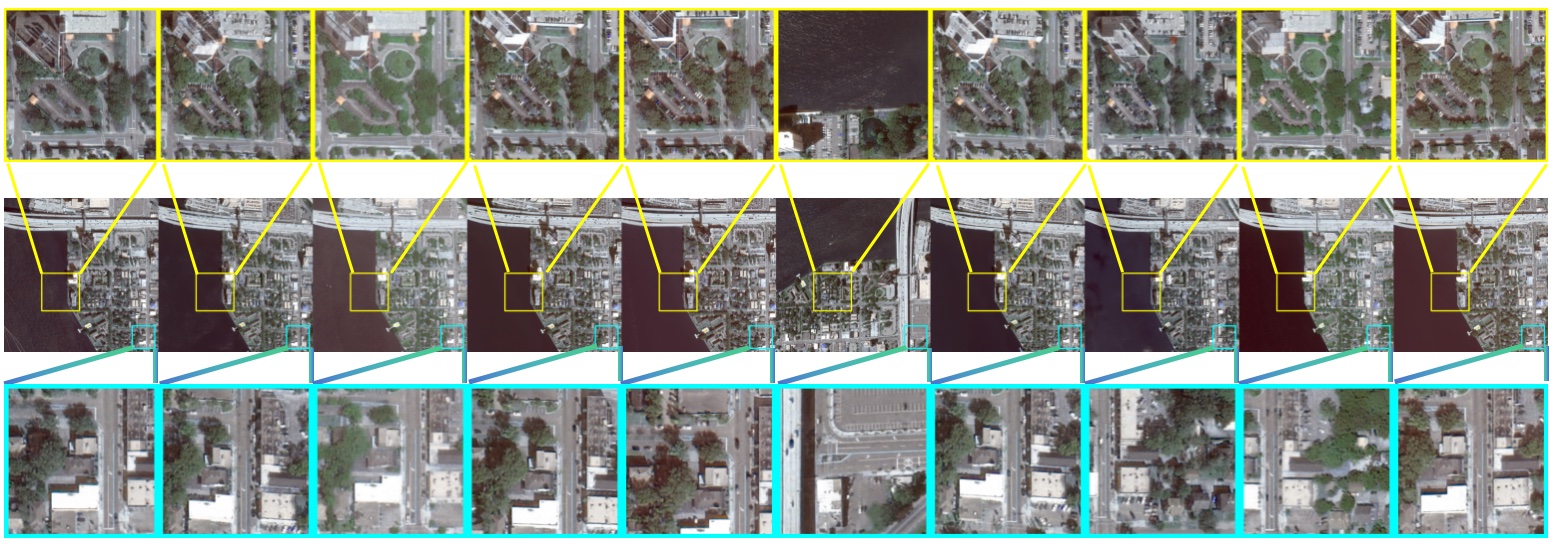}
    \caption{Details of the synthesized images in subset 260, with details of water body of size $500\times 500$ and  details of  vegetations of size $300\times 300$.}
    \label{fig:260}
\end{figure*}
\begin{figure*}
          \centering
    \includegraphics[width=18cm]{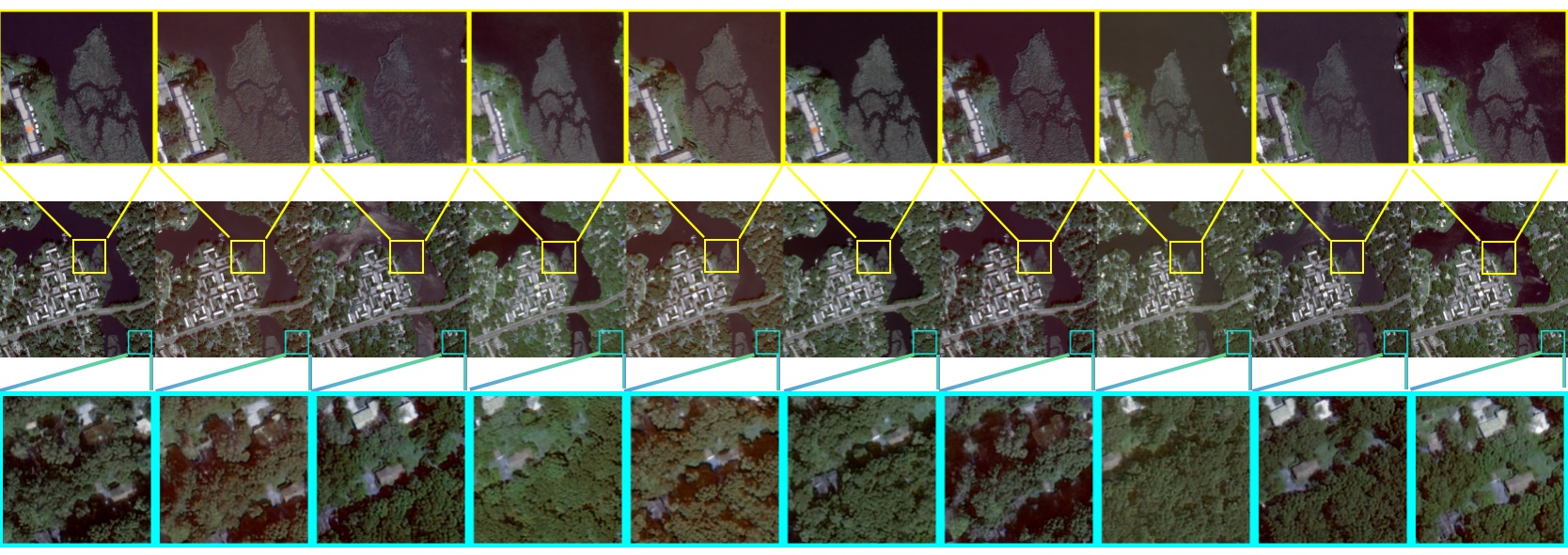}
    \caption{Details of the synthesized images in subset 280, with details of water body of size $500\times 500$ and  details of  vegetations of size $300\times 300$.}
    \label{fig:280}
\end{figure*}

\subsubsection{Single-date Comparison}
In the same model zoo, for single-date imagery, Table \ref{tab:comp2sota_single} documents a comparative analysis between SatensoRF and state-of-the-art neural radiance field models. These evaluations were conducted on test and training images from a neighboring area. We stop trainings once the validation PSNR reaches a threshold of 37.0. Consequently, all models demonstrate comparable performance on the test set and differ a little on the training set.
It is worth noting that, with the same lightning conditions, the additional modules of S-NeRF and Sat-NeRF does not promote synthesis quality than NeRF, while vanilla NeRF is faster.  With similar performance, SatensoRF converges faster than all the NeRF-based models with the same level of performance. 
\begin{table*}
    \centering
        \caption{Numerical results comparisons of SatensoRF and the SOTA work on single date imagery. Postfix \textit{-SC} refers to models with solar correction supervision in Sat-NeRF paper \cite{satnerf}, and \textit{-DS} refers to models with depth supervision.}
    \begin{tabular}{c|cccc|cccc}  
    \toprule 
    
    &
    \multicolumn{4}{c}{\textbf{PSNR}$\uparrow$(test/train)}
    &\multicolumn{4}{c}{\textbf{SSIM}$\uparrow$(test/train)}
    \cr  
    \cmidrule{2-9}
    Area Name&017-6&033-6&072-6&251-6&017-6&033-6&072-6&251-6\cr 
    \midrule
     NeRF&  37.012/42.561&36.984/43.784 & 36.971/43.659&36.969/42.998&0.901/0.923&0.982/0.988&0.913/0.968&0.948/0.952\cr
    \midrule[1pt]
    S-NeRF + SC& 36.925/45.032&35.779/43.291&36.216/42.145&35.968/40.328&0.896/0.972&0.878/0.984&0.915/0.983 &0.904/0.936\cr
    \midrule[1pt]
    Sat-NeRF + SC &36.546/43.592&37.004/44.672&36.882/44.105&36.829/43.045&0.935/0.972&0.921/0.986&0.905/0.928&0.931/0.979 \cr
\midrule[1pt]
SatensoRF + SC&\textbf{36.942/42.306}&\textbf{37.031/43.694}&\textbf{36.699/43.320}&\textbf{36.829/42.984}&\textbf{0.955/0.988}&\textbf{0.932/0.976}&\textbf{0.959/0.992}&\textbf{0.946/0.984} \cr
    \bottomrule  
    \end{tabular}   
    \label{tab:comp2sota_single}
\end{table*}
 \begin{table*}
    \centering
        \caption{Numerical results comparisons of SatensoRF with BRDF assumptions and Lambertian assumption.}
    \begin{tabular}{c|ccccc}  
    \toprule 
    &\textbf{PSNR}$\uparrow$(test/train) & \textbf{SSIM}$\uparrow$(test/train) & \textbf{Altitude MAE}(m)$\downarrow$(test/train)& $T^{multi}_{train}(s)$ & $T^{single}_{train}(s)$
    \cr   
    \midrule

lamb CP + SH & 20.382/24.071& 0.792/0.827&-&\textbf{8529.31}&\textbf{57.27}\cr
\hline
non lamb CP + asg & 21.518/24.599& 0.781/0.847&-&10010.29&79.39\cr
\hline
lamb VM + SH& 25.473/30.074&0.872/0.901&3.104/2.945&10113.83&93.99 \cr
\hline
non lamb VM + SH&25.819/29.818&0.885/0.893&2.964/2.861&10138.73&95.85 \cr
\hline
non lamb VM + asg& \textbf{26.695/32.705}&\textbf{0.918/0.952}&\textbf{ 2.081/2.021}& 10129.81&97.49 \cr

    \bottomrule  
    \end{tabular}   
    \label{tab:ablation_lambtian}
\end{table*}
 \begin{table*}
    \centering
        \caption{Numerical results comparisons of SatensoRF with total variation regularization and L1 regularization.}
    \begin{tabular}{c|ccccc}  
    \toprule 
    Regularization&\textbf{PSNR}$\uparrow$(test/train) & \textbf{SSIM}$\uparrow$(test/train) & \textbf{Altitude MAE}(m)$\downarrow$(test/train)& $T^{multi}_{train}(s)$ & $T^{single}_{train}(s)$
    \cr   
    \midrule
TV& \textbf{26.695/32.705}&\textbf{0.918/0.952}& \textbf{2.081/2.021}& 10129.81&97.49 \cr
\hline
L1& 26.314/31.059&0.857/0.886&-&\textbf{10096.53}&\textbf{97.29} \cr
\hline
L1 + TV &26.518/31.622&0.872/0.894&-&10359.47&97.39\cr

    \bottomrule  
    \end{tabular}   
    \label{tab:ablation_tv}
\end{table*}
\begin{table*}
    \centering
        \caption{Performance Comparison by changing the tensor field level $L$.}
    \begin{tabular}{c|ccccccc}
    \toprule
         & \textbf{PSNR}(dB)$\uparrow$ &\textbf{SSIM} $\uparrow$&\textbf{Altitude MAE}$(m)\downarrow$ &$T_{train}(s)$ & $T_{test}(s)$ &$\#Params(M)$ & $FLOPs^{multi}(B)$\\
         \midrule
         layer-2 $\mathcal{T}_{\sigma}$-2&23.208  &0.877&5.382&\textbf{8189.29}&\textbf{57.23}&0.24&129.52$b$ \\
         \hline
         layer-2 $\mathcal{T}_{\sigma}$-4& 24.098 &0.894&4.367&9101.337&70.405&0.25&132.74$b$\\
 \hline
         $layer$-4$ \mathcal{T}_{\sigma}$-8& 27.031 &0.922&2.009&10452.052&105.31&0.86&450.57$b$\\
          \hline
         $layer$-2  $\mathcal{T}_{\sigma}$-8(SatensoRF)&  \textbf{26.695}&\textbf{0.918}&\textbf{2.081}&10129.813&97.495&0.79&416.21$b$\\
         \bottomrule
    \end{tabular}
    \label{tab:ablation_Ms}
\end{table*}
\begin{figure*}
          \centering
    \includegraphics[width=16cm]{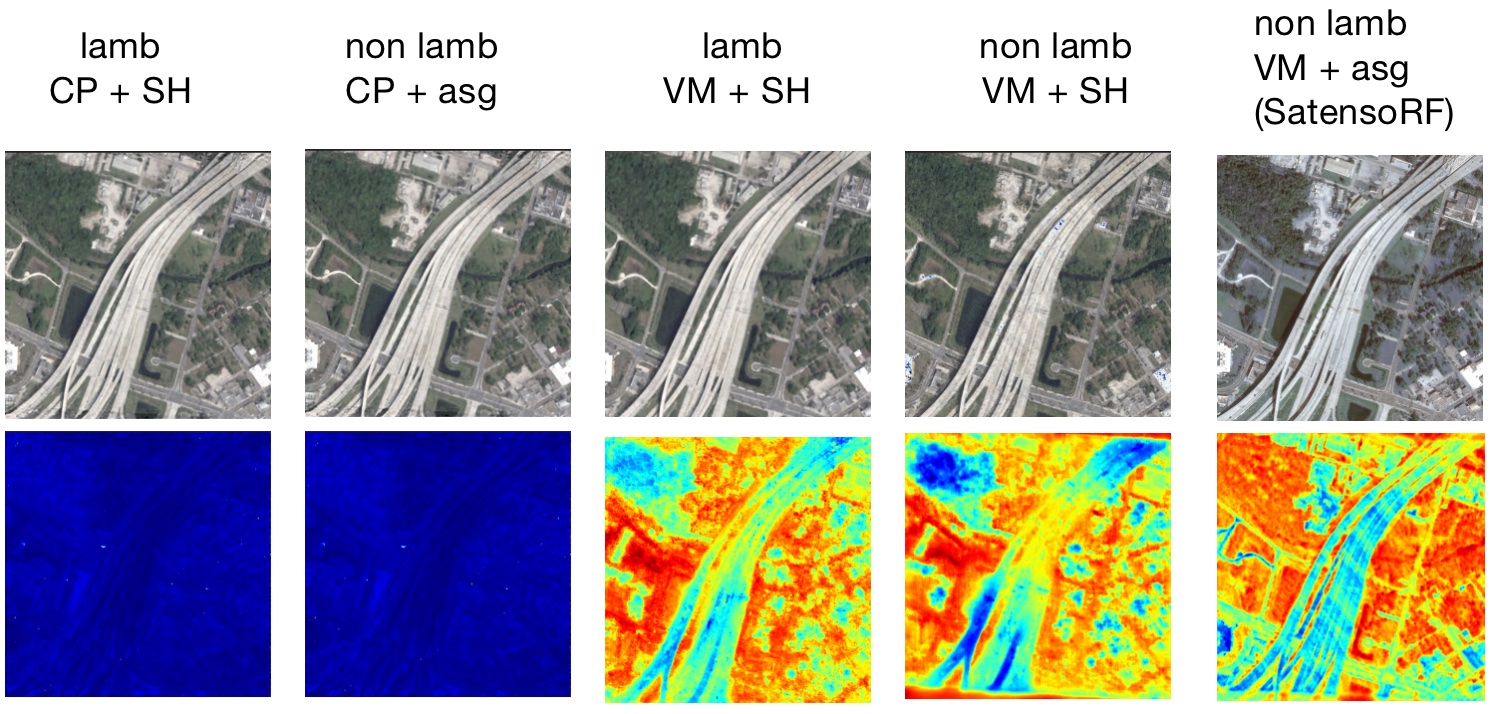}
    \caption{The comparison of whether using Lambertian Assumption.}
    \label{fig:ablation_lambtian}  
\end{figure*}


\subsection{Ablation Study I: Non Lambertian Assumption}
\label{sec:validate_nonlamber}
 This section aims to validate the non-Lambertian assumption by modeling the anisotropic reflectance field. The candidates for comparison are as follows:
\begin{itemize}
    \item lamb CP + SH: The Lambertian version  version of SatensoRF without modeling the reflective light. The network has only the diffusion head (MLPs of 4 layers with 128 hidden neurons), and no other output from module A in Fig.~\ref{fig:lightdecomp}.
    The final RGB output is rendered with spherical harmonics functions (SH), and the tensor factorization adopts CP (CANDECOMP/PARAFAC) decomposition. 
    \item non lamb CP + asg: The non Lambertian version of SatensoRF modeling the reflective light with CP decomposition.  The network is the same as module A in Fig.~\ref{fig:lightdecomp} and the final RGB is rendered with asg functions.  
    \item lamb VM + SH: The same with lamb CP + SH but the tensor radiance fields are composed of VM (Vector-Matrix) decomposition.
        \item non lamb VM + asg: The SatensoRF itself.
    \item non lamb VM + SH: The same with non lamb VM + asg, but the final RGB is rendered with SH functions.

\end{itemize}
We evaluate the performance of these models  on the test set and training set of subset 159.
The quantitative results of performance and the model are listed below in Table~\ref{tab:ablation_lambtian}, and the visualizations are summarized in Fig.~\ref{fig:ablation_lambtian}.
From Fig.~\ref{fig:ablation_lambtian} we can see that the replacing VM decomposition with CP decomposition results in wrong volume density computation, and the wrong altitude map can not produce DSM product correspondingly so MAE is ommited in Table~\ref{tab:ablation_lambtian}.

\subsection{Ablation Study II: the Total Variation Loss}
\label{sec:validate_tv}
This section seeks to validate the rationale behind using the total variation loss, to mitigate the adverse effects of transient objects as noise on the volume density field $\mathcal{T}_{\sigma}$. In the default configuration of SatensoRF, the loss function specified in Equation~\ref{eq:total_loss} is employed.

The comparative analysis encompasses the following factors:
\begin{itemize}
    \item TV: Training SatensoRF with  $\mathcal{L}_{TV}$.
    \item L1: Training SatensoRF with  $\mathcal{L}_{l1}$.
    \item L1 + TV: Training SatensoRF with both $\mathcal{L}_{l1}$ and $\mathcal{L}_{TV}$
    
\end{itemize}
where $\mathcal{L}_{l1}(x) = \frac{1}{M}\sum\limits_{i=1}^{M-1}|\mathcal{T}_{\sigma}|$.

In Table \ref{tab:ablation_tv},
the validity of total variation loss $\mathcal{L}_{TV}$ is significant since it preserves the right geometry of the area, while  $\mathcal{L}_{l1}(x)$ leads the model to the wrong volume density with nice rendering results. In 
Besides, Table \ref{tab:ablation_tv} shows that once $\mathcal{L}_{l1}(x)$ is adopted, whether $\mathcal{L}_{TV}$ involves can not guarantee the volume density estimation.

\subsection{Controlled Experiments}
\label{sec:controlled_exp}
 SatensoRF is comprised of tensor composited fields and MLPs, and in this study, we delve into how the design of these two components influences the overall performance and efficiency of the model.
For brevity, we denote model $\mathcal{T}_{\sigma}-m$ to be the SatensoRF model with $m$ level of tensor radiance fields; we denote model $layer-n$ to be the SatensoRF model with MLP network with $n$ layers. 

Table \ref{tab:ablation_Ms} shows that both increasing the levels of tensor radiance field or the network layers enhance the performance, but the enhancement is limited while the time and computation costs is unworthy.
Additionally, insufficient level of tensor radiance field causes poor result in 3D scene reconstruction.

\begin{figure}[t]
    \centering
    \includegraphics[width=8cm]{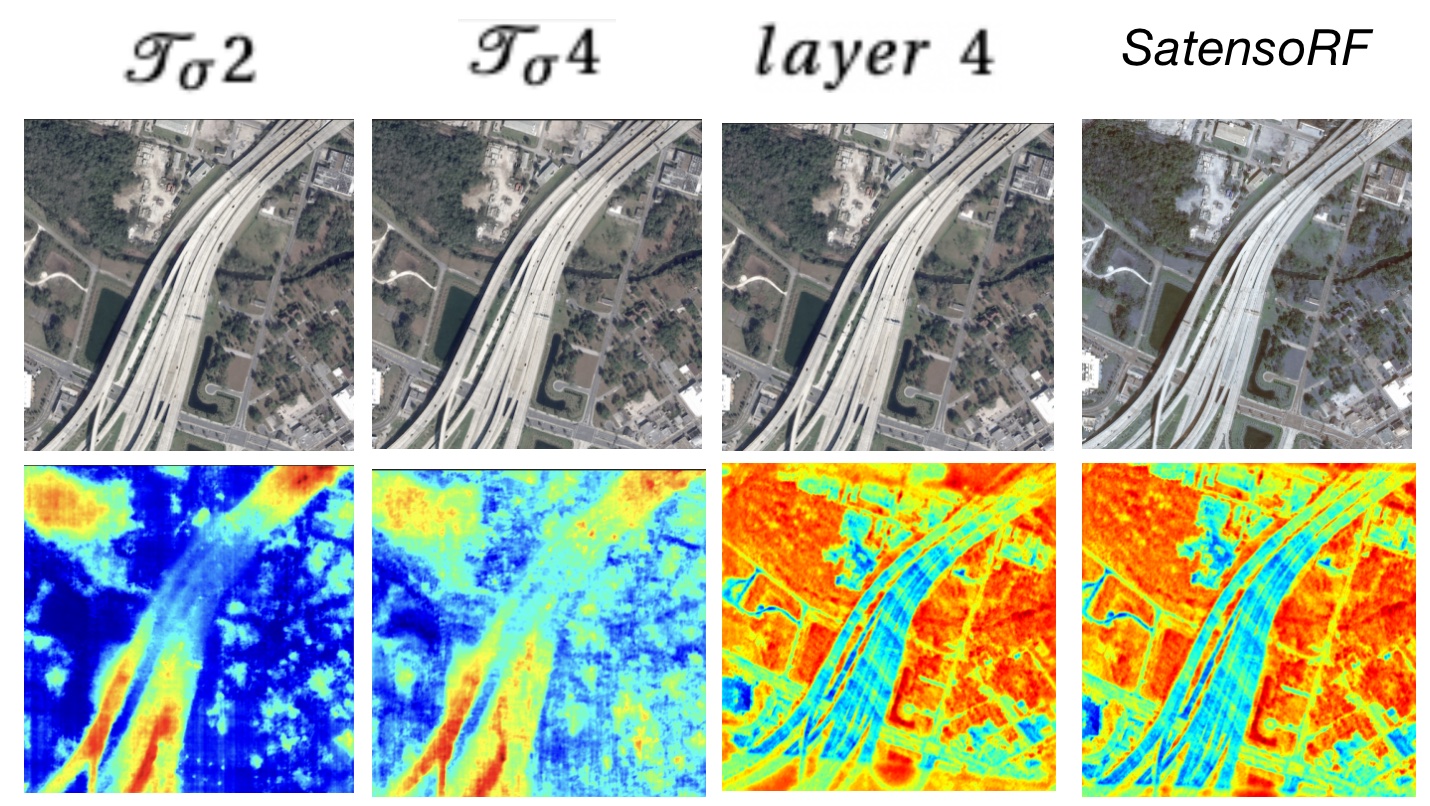}
    \caption{The result comparison of the controlled experiments}
    \label{fig:controlled_exp}
\end{figure}
\section{Conclusion}
We propose a fast, level-of-detail tensorized neural radiance field called SatensoRF for large scale satellite novel view synthesis. The proposed radiance field efficiently speeds up the training and inference process, with better rendering quality. At the level of principles, SatensoRF firstly proposes to model the anisotropic light field rather than assuming the earth surface as Lambertian, to boost the radiance field performance. Moreover, SatensoRF are free from extra input such as solar directions.
 We conduct extensive experiments to evaluate SatensoRF on both  performance and efficiency. SatensoRF shows better image rendering quality than SOTA Sat-NeRF series with much less computation, memory and training time on large satellite images. 
 For future work, this paradigm can be extended with new dimension, such as time, to enable long-time ground surface observation.

 \section*{Acknowledgment}
This research is supported in part by the Special Funds for Creative Research Grant (No. 2022C61540).
\bibliographystyle{IEEEtran}
\bibliography{ref}

\end{document}